\documentclass[conference]{IEEEtran}
\usepackage{cite}
\usepackage{amsmath,amssymb,amsfonts}
\usepackage{algorithmic}
\usepackage{graphicx}
\usepackage{textcomp}
\usepackage{xcolor}

\usepackage{multicol}
\usepackage{hyperref}

\def\BibTeX{{\rm B\kern-.05em{\sc i\kern-.025em b}\kern-.08em
		T\kern-.1667em\lower.7ex\hbox{E}\kern-.125emX}}

\title{Integrating Motion into Vision Models for Better Visual Prediction
}

\author{\IEEEauthorblockN{Michael Hazoglou}
\IEEEauthorblockA{\textit{Department of Electrical and Computer Engineering} \\
	\textit{University of California San Diego}\\
	San Diego, USA \\
	mhazoglou@eng.ucsd.edu}
\and
\IEEEauthorblockN{Todd Hylton}
\IEEEauthorblockA{\textit{Department of Electrical and Computer Engineering} \\
	\textit{University of California San Diego}\\
	San Diego, USA \\
	thylton@eng.ucsd.edu}
}

\begin{document}
	\maketitle
	
	\begin{abstract}
		We demonstrate an improved vision system that learns a model of its environment using a self-supervised, predictive learning method.  The system includes a pan-tilt camera, a foveated visual input, a saccading reflex to servo the foveated region to areas high prediction error, input frame transformation synced to the camera motion, and a recursive, hierachical machine learning technique based on the Predictive Vision Model.  In earlier work, which did not integrate camera motion into the vision model, prediction was impaired and camera movement suffered from undesired feedback effects.  Here we detail the integration of camera motion into the predictive learning system and show improved visual prediction and saccadic behavior.  From these experiences, we speculate on the integration of additional sensory and motor systems into self-supervised, predictive learning models.
		%
	\end{abstract}
	
	
	\section{Introduction}
	\label{sec:intro}
	Deep learning models are incredibly successful with specific tasks such as classification problems or achieving super-human skills playing in games \cite{DBLP:journals/corr/MnihHGK14,campbell2002deep,lewis2012game,silver2016mastering,2013arXiv1312.5602M}, but these models have well known limitations. For example, classification problems require large amounts of labeled data for supervised training of models and reinforcement learning requires a reward function associated with inputs and actions.  A classifier built to distinguish cars from airplanes is often fooled by a collage of features from cars or airplanes and by more subtle, engineered adversarial inputs. Reinforcement learning based models are generally aware of the entire state of the game that they are learning to play.  In general these techniques rarely generalize to other similar tasks without using specialized training methods such as transfer learning or meta-learning and usually require more data for training. These limitations are particularly apparent when considering the challenge of creating an intelligent agent capable or interacting with the world. These agents will never have complete information about their environment and must build simplified models based on limited information that are relevant and general to the agent's needs or tasks. We suppose that future computing systems will need such capabilities more generally: that is, computing systems will need to spontaneously learn simplified, relevant models of their data environments with minimal, task-specific configuration and programming.  In this contribution, we illustrate such ideas in the context of a vision system that learns its environment by interacting with it.
	
	In previous work \cite{Hazoglou:2018:SPV:3229884.3229886} we developed a biologically inspired model to address some of these issues, by focusing on a simple problem of visually surveying an environment by emulating saccades; the rapid movements of the eye. This Error Saccade Model (ESM) uses the prediction error of the Predictive Vision Model (PVM) \cite{DBLP:journals/corr/PiekniewskiLPRF16} to determine where to move the field of view based on where the total error in a sliding window is maximal. The intuition behind the ESM is that our eyes are naturally attracted to unexpected events. Taking our biological inspiration further we modified the PVM to include a fovea, which is a pit with a high density of cones in the eye (cone cells are photo receptors responsible for detecting color). As compared to PVMs without foveas, we showed that the PVM with a fovea spent significantly more time with its field of vision on complicated (high image entropy) objects. In this article the authors propose incorporating the motion of the camera into the visual prediction by applying a simple linear transformation according to the corresponding movement.

	\section{The Predictive Vision Model}
\begin{figure}
		\centering
		\includegraphics[width=\columnwidth]{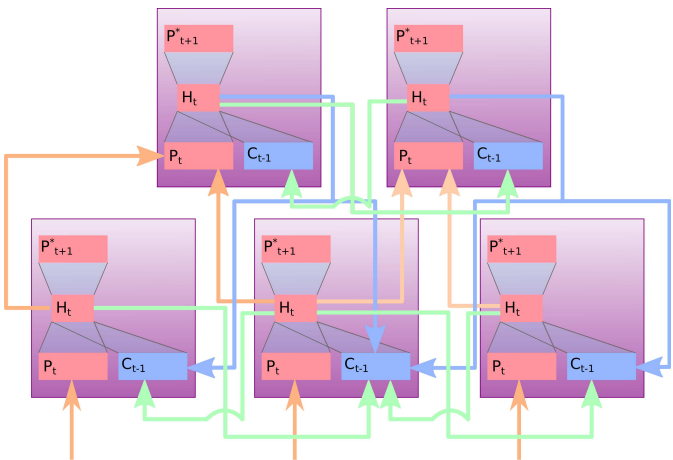}
		\caption{An example of the PVM hierarchy used in \cite{DBLP:journals/corr/PiekniewskiLPRF16}. The purple boxes represent the PVM units. Orange arrows represent the inputs into the units. The blue arrows show the flow of context from superior units while the green arrows show the flow of lateral context. The primary signal (bottom large red box) and context (in blue) is compressed to generate the (hidden) state of the unit which is shown as the smaller red box from which a prediction (top large red box) is made for the next input the unit will receive. Since the inputs to the units are fed one at a time, time derivatives, error in the previous prediction and time averages (integrals) are also factored in the hidden state calculation.}
		\label{BasePVMhierarchy}
	\end{figure}

	The Predictive Vision Model (PVM) is a self-supervised hierarchical recurrent neural network that takes an input sequence and predicts the next input. The hierarchy of the PVM is composed of individual units which compress their respective inputs and exchange their hidden contextual information among each other to produce a prediction. Individual units of the PVM perform the calculations listed in Table~\ref{PVMunitCalculation}. The hierarchy is split into different levels where the next higher adjacent level of the hierarchy predicts the hidden states of the lower level, and the higher level feeds its context down to this lower adjacent level (see Figure~\ref{BasePVMhierarchy}). The hierarchies are built from rectangular grids at each level with nearest neighbor connections. Between adjacent levels in the hierarchy connections are formed in pyramidal arrangements such that typically 4 units in square arrangement pass their context up to one unit in the superior level. This reduces the number of units at each level resulting in a pyramidal structure. The PVM is trained in an self-supervised manner using backpropagation with a squared error loss on each PVM unit's prediction of its future input signal. For a more detailed discussion of the Predictive Vision Model see \cite{DBLP:journals/corr/PiekniewskiLPRF16}. We reimplemented the PVM in previous work to run on Nvidia graphics cards see  \cite{Hazoglou:2018:SPV:3229884.3229886} for the detailed algorithm. All the code is available at \url{https://github.com/mhazoglou/PVM_PyCUDA} including scripts that were written for data collection and training.


	\begin{table}
		\begin{tabular}{| p{1.5cm} | c | p{5cm} |}
			\hline 
			Layer & Symbol & Definition \\ \hline
			\multicolumn{3}{|c|}{\textbf{Inputs}} \\ \hline
			Signal & $P_t$ & Fan-in from inferior level or raw video tile \\ \hline
			Integral & $I_t$ & $\tau I_{t-1} + (1-\tau)P_t$\\ \hline
			Derivative & $D_{t/t-1}$ & $0.5 + (P_t - P_{t-1})/2$ \\ \hline
			Previous Prediction Error & $E_t$ & $0.5 + (P^*_t - P_t)/2$ \\ \hline
			Context & $C_{t-1}$ & concat[$H_t \dots$] from Hidden of self/lateral/superior/topmost units \\ \hline
			\multicolumn{3}{| c |}{\textbf{Output}} \\ \hline
			Hidden & $H_t$ & $\sigma(W_h \cdot [P_t; D_{t/t-1};I_t;E_{t};C_t] + b_h) $ \\ \hline
			\multicolumn{3}{| c |}{\textbf{Predictions}} \\ \hline
			Predicted Signal & $P^*_{t+1}$ & $\sigma(W_p \cdot H_t + b_p) $ \\ \hline 
			\multicolumn{3}{| c |}{\textbf{Loss Function}} \\ \hline
			Square Error & $\mathcal{L}_{t}$ & $\sum_{\{\mbox{channels, pixels}\}} (P^*_{t+1} - P_{t+1})^2$ \\ \hline 
		\end{tabular}
		\caption{Summary of the PVM unit. Each unit consists of a three layer perceptron with sigmoid action neurons. The indices represent the time step for the respective value. For more details see \cite{DBLP:journals/corr/PiekniewskiLPRF16}.}
		\label{PVMunitCalculation}
	\end{table}
		
	\section{The Error Saccade Model and Fovea}
	\label{sec:ESMandRovea}
	The fovea-like structure is created by splitting units of the PVM in the central region of lowest level into smaller and more numerous units. As the PVM units are split they remain connected to the units to which they were originally connected.
	
	The error saccade model utilizes forced damped harmonic oscillation with gaussian white noise to emulate the movements of saccades. The total squared error in a sliding window of fixed height and width is calculated across the prediction error of the input level of the PVM hierarchy, the window with the largest error is used as the equilibrium point if the total squared error is larger than a threshold value. The threshold is equal to the time average of the maximum total square error in these sliding windows. This means that as the predictions get better/worse the movement will be more/less sensitive to prediction error. The error saccade model is complemented by the foveal region as the effect of having a lower density of PVM units in the periphery of the field of view results in higher prediction error, which induces saccades to that point. The foveal region has a higher density of PVM units meaning the prediction will have more detail which is beneficial for detailed (high image entropy) scenes. This means that a detailed object will likely stay in the fovea as any drift to the peripheral lower density region will cause a spike in error which will recenter fovea on the object. As we observed in our previous work, the model replicates certain features of saccades and smooth pursuit eye movements observed in humans \cite{yarbus1967eye,Hazoglou:2018:SPV:3229884.3229886} as the models with the fovea spent more time in regions with higher image entropy than models with uniform densities.
	
	\section{Data Collection}
	A Playstation Eye web camera mounted on a PhantomX turret from Trossen Robotics with Dynamixel AX-12A robot actuators was used in the work described here. To obtain data a controller reads the pose of the servos and writes the values to a serial connections when pinged. Videos were collected from the camera at a resolution of 320 by 240 pixels with a frame rate of 120 frames per second using a buffer size of a single frame. Using a buffer size of one frame is key for synchronizing the servo poses and frames, as only the most recent frame is needed when the controller pings for the pan-tilt pose. Data were collected as sets of a thousand frames and corresponding servo positions; fifty sets (50,000 frames) were used for model training and fifty different sets were used for testing. For each data set, the turret motion was randomly selected from a set of twenty pan and twenty tilt predefined trajectories (four hundred possible trajectories in total) involving oscillation, reflective motion and resetting position, running at different rates and in reverse. Each frame of the video was scaled down to 128 by 96 pixels for training and testing. All video was recorded using the wider field of view setting of 75$^{\circ}$ on the camera.
	
	\section{Incorporating Motion into the Prediction}
	The work on the error saccade model and fovea described in section~\ref{sec:ESMandRovea} motivated us to create a physical demonstration using the camera and turret previously described. We quickly learned that the movement of the camera induced errors in the prediction from the PVM that would influence its own motion, causing it to very frequently get caught in fixed trajectory loops. With the turret having an unloaded rotation speed of 300$^{\circ}$/sec and the phenomenon of saccades we were emulating achieving speeds in excess of that (~1000$^\circ$/sec), we knew that integrating the effect of the motion into the visual prediction was necessary to prevent the endless loops and improve overall performance.
	
	We decided to use a high frame rate (120 frames/sec) sufficient to remove motion blur and to enable us to focus on the effects of perspective changes caused by the movement of the camera. In general, the turret creates both displacement and rotation of the camera, but in the model described here we ignore the effects of camera displacement.  By ignoring displacement in camera position we also ignore any effects of occlusion as the camera moves, such as moving beyond a corner and seeing something behind it. With this approximation, the change in perspective can be described as a rotation in three dimensions. We choose the $x$ and $y$ directions in the image plane (positive $y$-axis being oriented down the vertical with positive $x$-axis directed right along horizontal direction), and the $z$-direction along the center line of the camera's field of view. The transformation $T(\phi_2, \theta_2; \phi_1, \theta_1)$ can be expressed in terms of Euler angles depending only on the initial pan and tilt angles $\theta_1$ and $\phi_1$, respectively and the final pose's pan and tilt angles $\theta_2$ and $\phi_2$. 
	\begin{multline}
	    T(\theta_2, \phi_2; \theta_1, \phi_1) = \\
	    R_x(\phi_2)R_y(\theta_2)R_y^{-1}(\theta_1)R_x^{-1}(\phi_1) = \\
	    R_x(\phi_2)R_y(\theta_2 - \theta_1)R_x^{-1}(\phi_1)
	    \label{Transform}
	\end{multline}
	Where $R_x$ and $R_y$ denote a rotation about the $x$ or $y$ axis respectively. Explicitly these rotation matrices are written as 
	
	\begin{equation}
	    R_x(\alpha) = \begin{pmatrix}
	        1 & 0 & 0 \\
	        0 & \cos(\alpha) & \sin(\alpha) \\
	        0 & -\sin(\alpha) & \cos(\alpha) 
	    \end{pmatrix}
	\end{equation}
	
	\begin{equation}
	    R_y(\alpha) = \begin{pmatrix}
	        \cos(\alpha) & 0 & \sin(\alpha) \\
	        0 & 1 & 0 \\
	        -\sin(\alpha) & 0  & \cos(\alpha) 
	    \end{pmatrix}
	\end{equation}
	
	Eq.~\eqref{Transform} can be written as
	
	\begin{equation}
	    \begin{pmatrix}
	        \cos\theta_{21} & \begin{matrix} \sin\phi_1 \sin\theta_{21} & -\cos\phi_1\sin\theta_{21} \end{matrix} \\
	        \begin{matrix}
	            -\sin\phi_2 \sin\theta_{21} \\
	            \cos\phi_2\sin\theta_{21} 
	        \end{matrix} & Q
	    \end{pmatrix}
	\end{equation}
	
	\begin{multline}
	    Q = \cos\theta_{21}
	    \begin{pmatrix}
	       \sin\phi_2\sin\phi_1 &  -\sin\phi_2\cos\phi_1 \\
	       -\sin\phi_1\cos\phi_2  & \cos\phi_1\cos\phi_2 
	    \end{pmatrix} \\
	    +
	    \begin{pmatrix}
	        \cos\phi_2 \cos\phi_1 & \cos\phi_2\sin\phi_1 \\
	        \sin\phi_2\cos\phi_1 & \sin\phi_2\sin\phi_1
	    \end{pmatrix}
	\end{multline}
	with $\theta_{21} = \theta_2 - \theta_1$. This gives us the affine transformation on the horizontal and vertical positions (after rescaling by the focal length in it's pixel equivalent) of each pixel in the input image by applying the transformation in Eq.~\eqref{Transform} gives,
	
	\begin{equation}
	\begin{pmatrix}
	    x' \\
	    y' \\
	    z'
	\end{pmatrix} 
	= T(\theta_2, \phi_2; \theta_1, \phi_1)
	    \begin{pmatrix}
	        x \\
	        y \\
	        1
	    \end{pmatrix}.
	    \label{AffineTransform}
	\end{equation}
	As the mapping does not map directly onto pixel positions, interpolation can be used to find the appropriate pixel value or the pixel position can be rounded instead. The later method was used instead of interpolation, which may cause the transformed image sent to the PVM and the respective prediction to looked as if it is aliased. As the original image is translated, rotated and rescaled some pixels are not mapped from the input to the transform image leaving it incomplete. These blank pixels are filled by extending the edges of the original image such that they fill the entire frame.
	
	\section{Results and Discussion}
\begin{figure}
	    \centering
	    \includegraphics[width=\columnwidth]{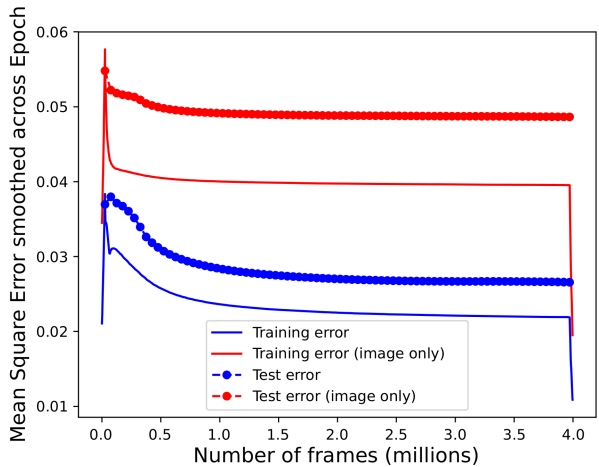}
	    \caption{Training and testing error for the PVM trained without any motion integration. The training error is smoothed over an epoch's worth of frames with zero padding (50,000 frames) for comparison with mean testing error after that epoch. The values in red are the mean squared error for the prediction of the image only, in other words the prediction error of just the lowest level of the PVM hierarchy. The values in blue are the mean squared error for all levels of the hierarchy.}
	    \label{fig:TrainingNoMotionIntegration}
	\end{figure}

\begin{figure}
	    \centering
	    \includegraphics[width=\columnwidth]{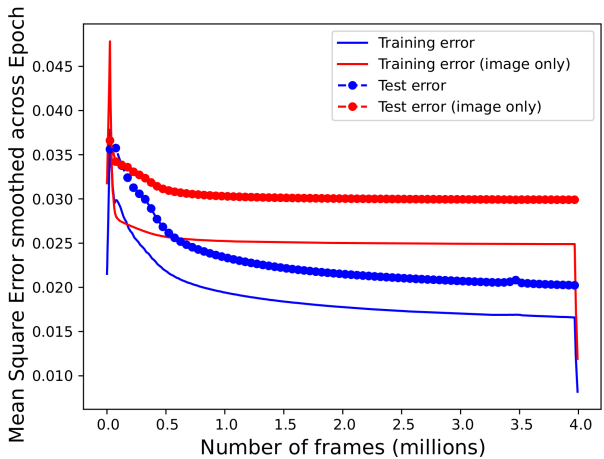}
	    \caption{Training and testing error for the PVM train with motion integration. The training error is smoothed over an epoch's worth of frames with zero padding (50,000 frames) for comparison with mean testing error after that epoch. The values in red are the mean squared error for the prediction of the image only, in other words the prediction error of just the lowest level of the PVM hierarchy. The values in blue are the mean squared error for all levels of the hierarchy.}
	    \label{fig:TrainingMotionIntegration}
	\end{figure}

	Two identically structured PVMs were constructed to compare the effect of integrating motion on the PVM. The structure is an eight level hierarchy of rectangular grids of PVM units with size 64 by 48, 32 by 24, 16 by 12, 8 by 6, 4 by 3, 3 by 2, 2 by 1 and 1 by 1 with each unit having a hidden size of 5 and the top most unit sending context to all lower units. The input into the model is a 128 by 96 pixel video sequence where 2 by 2 tiles of each frame are fed into to the corresponding unit in the lowest level of the PVM hierarchy. The mean square error versus frame count in the PVM without any motion integration is shown in Fig.~\ref{fig:TrainingNoMotionIntegration} and the PVM with motion integration is shown in Fig.~\ref{fig:TrainingMotionIntegration} for both the training and testing sets smoothed across a window of one epoch (50,000 frames).  The beginning and end of the training data sets are zero padded with 25,000 frames to maintain the total frame count (resulting in artifacts at the beginning and end of the training error curves in Fig.~\ref{fig:TrainingNoMotionIntegration} and Fig.~\ref{fig:TrainingMotionIntegration}). Comparing the two models shows that motion integration substantially improves the PVM's ability to predict the next video frame. Qualitatively, predicted image features are sharper and colors are less faded with motion integration than without it. Improved performance may also be due to permutation of noise due to the transformation \cite{ThangAndMatsui2019}. Both models struggle with predictions when very fast movements are involved as even the model with motion integration has no way to predict what lies outside it's field of view. Motion integration improving prediction should not be unique to just the PVM it should improve any other self-supervised visual prediction model \cite{Prednet2016,CortexNet2017}. 
	
	We applied the error saccade model with the PVM on the turret in a demonstration to compare the PVM with and without motion integration. One of the main issues with the behavior of the turret movement is that it can get caught in an infinite loop on a static background as a rapid back and forth between two fixation points is possible due to the induced errors caused by the movement of the camera. This is demonstrated in this video \url{https://bit.ly/2QbEfgs} where the turret keeps saccading between the two tables inducing high prediction errors. When motion is integrated into the model prediction the movement is more robust to endless loops, as seen in \url{https://bit.ly/2Wg0w20}. 
	
	\section{Conclusion}
	Here we have demonstrated an improved close-loop vision system that interacts and models its environment in an self-supervised fashion. The addition of motion integration substantially improved the PVM's prediction ability and when combined with the error saccade model does not have the pathology of being caught in endless loops. We speculate that stereoscopic vision with vergence movements would have the advantage of being able to judge distances and improve the predictive ability of a vision model especially when movement and occlusion are relevant as a three dimensional representation of a scene would be more informative than a representation from a single two dimensional image.
	
	\bibliographystyle{IEEEtran}
	\bibliography{ICRC2019}

\end{document}